%% file: acl_latex.tex
\documentclass[11pt]{article}

\usepackage[preprint]{acl}

\usepackage{times}
\usepackage{latexsym}

\usepackage{algorithm}
\usepackage{algorithmic}

\usepackage[T1]{fontenc}
\usepackage[utf8]{inputenc}

\usepackage{microtype}
\usepackage{inconsolata}
\usepackage{graphicx}

\usepackage{multirow}
\usepackage{booktabs}
\usepackage{float}
\usepackage{subcaption}
\usepackage{amsmath}
\usepackage{amssymb}
\usepackage{mathtools}
\usepackage{amsthm}
\usepackage[capitalize,noabbrev]{cleveref}
\usepackage[textsize=tiny]{todonotes}

\theoremstyle{plain}

\theoremstyle{definition}

\theoremstyle{remark}

\title{Robust Reasoning via Dynamic Token Selection \\for Distribution-Aligned Self-Distillation}

\author{
  \textbf{Ruiqi Zhang\textsuperscript{1,2,*}},
  \textbf{Lingxiang Wang\textsuperscript{1,2}},
  \textbf{Hainan Zhang\textsuperscript{1,2}},
  \textbf{Zhiming Zheng\textsuperscript{1,2}}
\\
\\
  \textsuperscript{1} Beijing Advanced Innovation Center for Future Blockchain and Privacy Computing, Beihang University \\
  \textsuperscript{2} School of Artificial Intelligence, Beihang University \\
\\
  \small{
    \textbf{Correspondence:} \href{mailto:zhanghainan@buaa.edu.cn}{zhanghainan@buaa.edu.cn}
  }
}

\begin{document}
\maketitle

\begin{abstract}

Self-distillation improves learning efficiency by rewriting reference answers as training data that better matches the model’s own distribution. However, reference answers also introduce strong stylistic biases, causing the generative model to imitate surface forms rather than learn useful reasoning patterns. We observe that the rewriting data contains a large number of high-perplexity(PPL) tokens, coming from two distinct sources: beneficial knowledge-enhancing logical corrections, and harmful stylistic drift induced by reference imitation. Treating all such tokens equally can disrupt the base model’s original distribution and degrade performance, especially on difficult reasoning tasks. To address this, we propose Distribution-Aligned Self-Distillation (DASD), which uses an answer-aware reference model to generate candidate tokens and dynamically filters them according to the base model’s confidence. DASD preserves tokens that encode useful logical knowledge while suppressing distributionally misaligned style noise. Experiments on math, code, and commonsense reasoning benchmarks show that DASD consistently outperforms competitive baselines, reduces high-PPL tokens, and improves robustness across tasks of varying difficulty.
\end{abstract}


\input{paper_content/1introduction}

\input{paper_content/2relatedwork}
\input{paper_content/3motivation}
\input{paper_content/4method}

\input{paper_content/5experiment}

\input{paper_content/6conclusion}

\bibliography{main}

\appendix
\section{Prompt Templates of Base and Reference Model}
\label{sec:appendix_prompt}

\begin{figure*}[t!]
    \centering
    \includegraphics[width=0.9\linewidth]{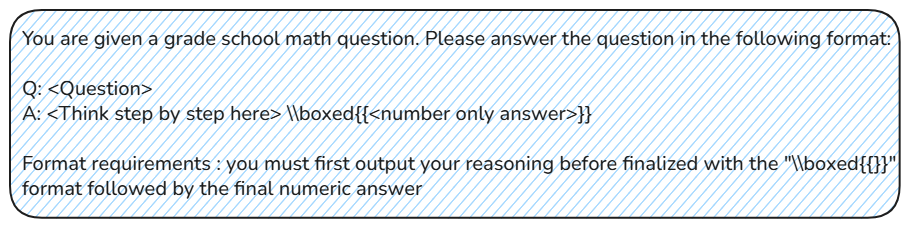}
    \caption{Prompt template of the base model}
    \label{fig:base_prompt}
\end{figure*}

\begin{figure*}[t!]
    \centering
    \includegraphics[width=0.9\linewidth]{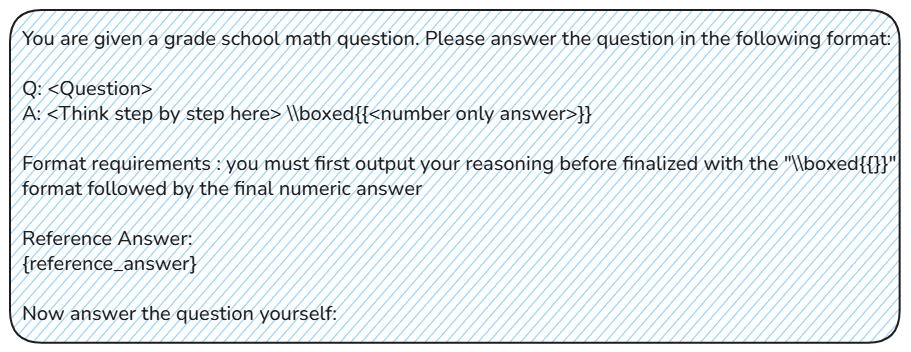}
    \caption{Prompt template of the reference model}
    \label{fig:ref_prompt}
\end{figure*}

We detail the prompt templates used for the base model and the reference model throughout our experiments. The base model takes only the raw problem question as input, following a standard step-by-step reasoning format without any additional reference information.
By contrast, the reference model is fed with the same question plus an official reference solution, which provides rigorous logical derivation and correct reasoning demonstrations.

The concrete prompt formats are illustrated in Figure~\ref{fig:base_prompt} and Figure~\ref{fig:ref_prompt}.
Figure~\ref{fig:base_prompt} presents the prompt template for the base model, which only contains the problem statement and format requirements.
Figure~\ref{fig:ref_prompt} shows the template for the reference model, where an extra reference answer is inserted to provide high-quality reasoning guidance.

This design enables the reference model to produce logically correct token candidates, while the base model preserves its inherent linguistic style and expression preference, supporting the token-level selection and distribution alignment in our method.

\section{Case Study}
\label{sec:appendix_case_study}

\begin{figure}[t]
    \centering
    \includegraphics[width=0.95\linewidth]{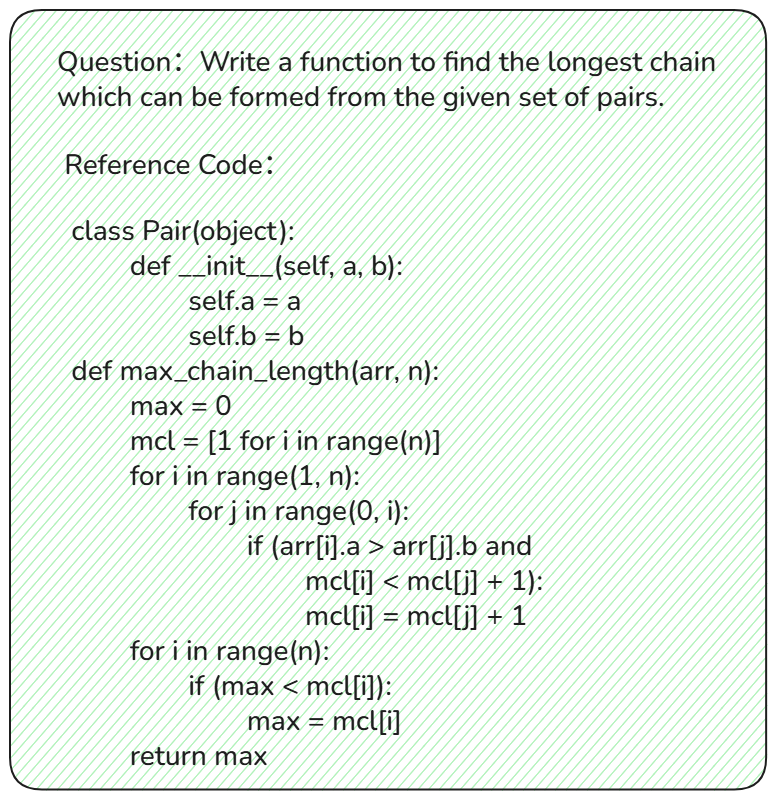}
    \caption{Original question and standard reference answer.}
    \label{fig:case_origin}
\end{figure}

\begin{figure}[t]
    \centering
    \includegraphics[width=0.9\linewidth]{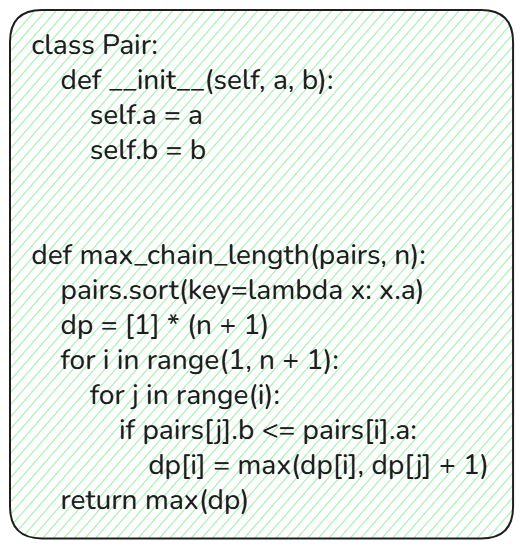}
    \caption{Generation output of the base model.}
    \label{fig:case_base}
\end{figure}

\begin{figure}[t!]
    \centering
    \includegraphics[width=0.9\linewidth]{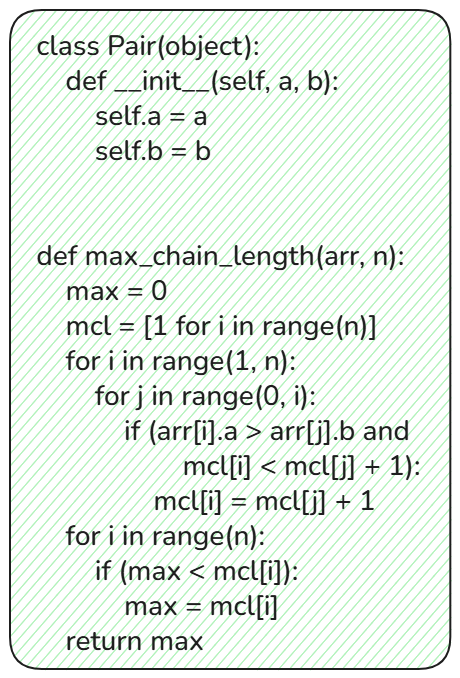}
    \caption{Generation output of the reference model. }
    \label{fig:case_ref}
\end{figure}

\begin{figure}[t!]
    \centering
    \includegraphics[width=1\linewidth]{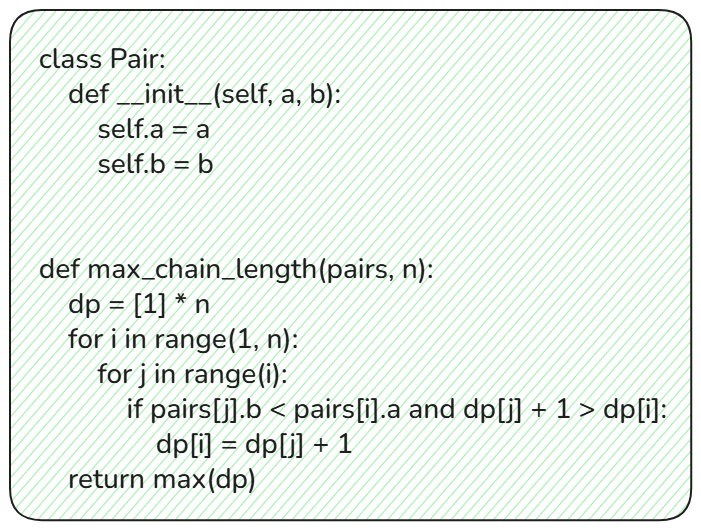}
    \caption{Generation output of our DASD method.}
    \label{fig:case_dasd}
\end{figure}

To intuitively compare the generation behaviors of different methods on code datasets, we present a representative case from the MBPP dataset. We visualize the original programming problem, the reference ground-truth solution, and the code responses generated by the baseline model, the reference model, and our proposed DASD method. The generated examples are shown in Figure~\ref{fig:case_origin}, Figure~\ref{fig:case_base}, Figure~\ref{fig:case_ref}, and Figure~\ref{fig:case_dasd}.

This case focuses on the classic Maximum Length of Pair Chain dynamic programming problem, which requires constructing the longest chain of pairs where the second element of each preceding pair is strictly less than the first element of the next pair. As illustrated in the generated code implementations, the baseline and reference models adopt standard dynamic programming paradigms with $O(n^2)$ time complexity: they initialize a DP array to record the longest chain length ending at each pair, iterate through all previous pairs to update the DP values, and finally return the maximum value of the DP array.

Notably, our DASD method preserves the natural output style and structural conventions of the base model, while implicitly injecting correctness signals derived from verified solutions. Compared to the baseline and reference responses, the DASD-generated code maintains consistency in variable naming, loop structure, and overall formatting, yet resolves subtle functional errors and logical oversights (e.g., off-by-one indexing issues, incorrect conditional bounds, and redundant array initialization) that appear in the baseline outputs. This demonstrates that DASD effectively enhances functional correctness without compromising the fluency and stylistic coherence of the original model’s generations.

\section{Supplementary Experiments}
Table~\ref{tab:llama8b_results} presents experimental results of Llama3-8B on MATH and commonsense reasoning tasks. The proposed DASD method still yields the best overall performance among all comparison approaches. On the MATH dataset, DASD reaches an accuracy of 0.292, outperforming the original baseline and other alternative methods. Conventional self-distillation fails to boost reasoning capability and even suffers performance drop, revealing that blindly learning from reference content easily breaks inherent token distribution. By contrast, our distribution-aligned self-distillation strategy avoids negative distribution drift and obtains favorable gains on both mathematical and commonsense reasoning scenarios.

\begin{table}[t!]
\centering
\caption{Results of Llama3-8B. \textbf{Bold} values denote the best performance, and \underline{underlined} values represent the second-best result.}
\label{tab:llama8b_results}
\renewcommand{\arraystretch}{1.10}
\small
\begin{tabular*}{0.5\textwidth}{@{\extracolsep{\fill}} lcc}
\toprule
Method & MATH & Commonsense Reasoning \\
\midrule
Base & \underline{0.278} & 0.799 \\
Self-Distillation & 0.248 & \textbf{0.819} \\
Mask-PPL & 0.258 & 0.810 \\
Hint-decoding & 0.264 & 0.814 \\
DASD & \textbf{0.292} & \underline{0.816} \\
\bottomrule
\end{tabular*}
\end{table}

\section{Efficiency Analysis}
Since the proposed DASD framework requires dual independent forward inferences from both the base model and the reference model, we simultaneously load and activate two model instances during training. This inevitably doubles the GPU memory consumption compared with conventional self-distillation paradigms. However, the additional token screening and confidence verification modules introduced in DASD are extremely lightweight. Their computational overhead is negligible relative to the full model forward propagation, resulting in nearly identical inference latency and no extra time cost for the overall data generation pipeline.

\end{document}

%% file: paper_content/1introduction.tex
\section{Introduction}
Catastrophic forgetting~\cite{luo2025empirical,li2024revisiting} remains a central challenge in post-training large language models(LLMs). When downstream data differs substantially from the pretraining distribution, direct fine-tuning can distort the model’s original parameter space. Self-distillation~\cite{yang2024self} mitigate this issue by transforming downstream examples into training samples that better match the model’s own generation distribution, thereby reducing distribution shift while improving training efficiency.

\begin{figure}[!t]
    \centering
    \includegraphics[width=0.4\textwidth]{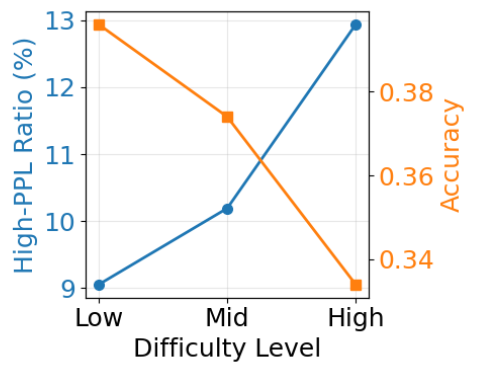}
    \caption{Correlation between average high-PPL token(PPL > 2.5) ratio and model performance under different difficulty levels. The x-axis denotes Low (level 1\&2), Mid (level 3\&4) and High (level 5).}
    \label{fig:moti1}
\end{figure}


However, self-distillation is not always distribution-aligned. In many downstream tasks, especially difficult reasoning problems, generated solutions are constrained to preserve ground-truth answers or reference reasoning traces. When the base model lacks sufficient reasoning competence, this process can cause the generated data to inherit reference-specific reasoning patterns, templates, and stylistic expressions rather than reflecting the base model’s native distribution. As shown in Figure~\ref{fig:moti1}, our hierarchical difficulty analysis on MATH benchmarks~\cite{hendrycks2021measuring} reveals that, as problem difficulty increases, the proportion of tokens inconsistent with the base model rises sharply, while the effectiveness of fine-tuning steadily deteriorates. This suggests that self-distillation may provide supervision that is correct in content but misaligned in distribution, weakening knowledge transfer and increasing the risk of forgetting.

A token-level analysis further shows that such inconsistencies arise from two qualitatively different sources, as shown in Figure~\ref{fig:framework}. Some high-perplexity(PPL) tokens represent beneficial logical corrections: reasoning steps that are unlikely under the base model but necessary for solving the problem. Others reflect harmful stylistic drift: redundant surface forms, reasoning templates, or answer-specific expressions inherited from the reference solution that contribute little useful knowledge. Naive self-distillation optimizes all generated tokens uniformly and therefore cannot distinguish useful corrections from noisy stylistic deviations.

Existing token-selection strategies only partially address this issue. Masked-PPL~\cite{wu2026mitigating} and ProFit~\cite{liu2026profit} methods remove high-PPL tokens from training, which suppresses stylistic noise but also discards valuable reasoning corrections. Hint-decoding~\cite{zhang2026towards} methods combine base and reference distributions using reference-model uncertainty, but answer-conditioned references can be overconfident, causing generation to remain biased toward the reference solution. These limitations call for a more selective mechanism that preserves useful reasoning deviations while filtering distributionally misaligned noise.

In this paper, we propose DASD, a confidence-based dynamic token selection method for robust Distribution-Aligned Self-Distillation. DASD constructs reference model’s candidate tokens to preserve answer correctness, while using the base model’s confidence to determine whether each token is distributionally acceptable. Tokens that are locally familiar to the base model are preferred to maintain distribution alignment, and a mandatory fallback mechanism retains indispensable reasoning tokens even when they are unlikely under the base model. In this way, DASD preserves beneficial logical corrections while suppressing harmful stylistic drift.

Experiments on mathematical reasoning, code generation, and commonsense reasoning benchmarks show that DASD consistently outperforms competitive self-distillation baselines. It reduces high-PPL stylistic deviations, improves robustness across difficulty levels, and better preserves pretrained knowledge during post-training~\footnote{https://anonymous.4open.science/r/emnlp-SD-6AB2.}. We summarize our main contributions as follows:
\vspace{-0.5em}
\begin{itemize}
    \item We expose self-distillation’s severe dependence on reference answers and verify how task difficulty and distribution shift affect performance.
    \item We propose DASD, a distribution-aligned self-distillation framework with confidence-aware dynamic token selection that balances knowledge injection and distribution preservation.
    \item We evaluate DASD across diverse reasoning benchmarks and show it substantially outperforms conventional self-distillation baselines, especially on difficult reasoning tasks.
\end{itemize}

%% file: paper_content/2relatedwork.tex
\section{Related Work}

\paragraph{Distribution-Consistent Data Selection}
Extensive efforts have explored training large language models with synthetic self-generated data~\cite{wang2023self}. In synthetic data construction, distribution alignment between training samples and the target model is widely recognized as critical, since models learn more efficiently on familiar data while mitigating the erosion of pre-trained knowledge~\cite{ren2024learn}.
To maintain stylistic consistency in model generation, SCAR~\cite{li2025scar} filters high-quality instruction data via style-aware ranking, which stabilizes model performance with fewer training samples. Beyond standard perplexity, the self-aligned PPL~\cite{ren2025efficiently} metric is proposed to measure the consistency between generated content and the model’s inherent reasoning patterns, facilitating better data selection. Methods such as Mask-PPL~\cite{wu2026mitigating} and ProFit~\cite{liu2026profit} directly discard high-perplexity tokens to preserve in-distribution content, yet they suffer from insufficient knowledge injection and limit the model’s ability to acquire new reasoning capabilities.


\paragraph{Self-Distillation}
Building on this, self-training with model-generated data has been extensively studied. Self-distillation~\cite{yang2024self} leverages the inherent distribution consistency of synthetic data to effectively alleviate distribution shift during fine-tuning.
Beyond Human Data~\cite{singh2023beyond} generates samples from the model, filters them via binary feedback, fine-tunes on correct samples, and repeats this process. In code generation tasks, SSD~\cite{zhang2026embarrassingly} stably improves model performance through balancing format paradigms and logical reasoning, even without strict correctness filtering on self-generated data. To ensure additional knowledge injection, STAR~\cite{zelikman2024star} prompts the model to regenerate answers by providing correct ones when it errs. The Self-Distillation work introduces reference answers with reasoning paths to assist model generation, ensuring both answer correctness and distribution proximity. Subsequent Hint-Decoding~\cite{zhang2026towards} research attempts to distinguish in-distribution and out-of-distribution content at the token level: it fuses outputs of the base and reference models via entropy, yet outputs are easily disturbed by reference answers, leading to inflated confidence and failure in accurately distinguishing style and logical tokens.

%% file: paper_content/3motivation.tex
\section{Motivation}
\label{sec:motivation}
Self-distillation adopts reference answers as external guidance to ensure the correctness of generated content. However, this mechanism causes the model to mechanically imitate reference-specific reasoning paths, leading to the generated content deviating from its original distribution. A typical phenomenon is the widespread emergence of high-perplexity tokens during self-generation. To reveal its adverse effects, we conduct a difficulty-stratified experiment on the MATH dataset~\cite{hendrycks2021measuring}. The results show that the higher the task difficulty, the higher the proportion of corresponding high-PPL tokens, and the further the model performance degrades.

\begin{figure}[!t]
    \centering
    \includegraphics[width=0.45\textwidth]{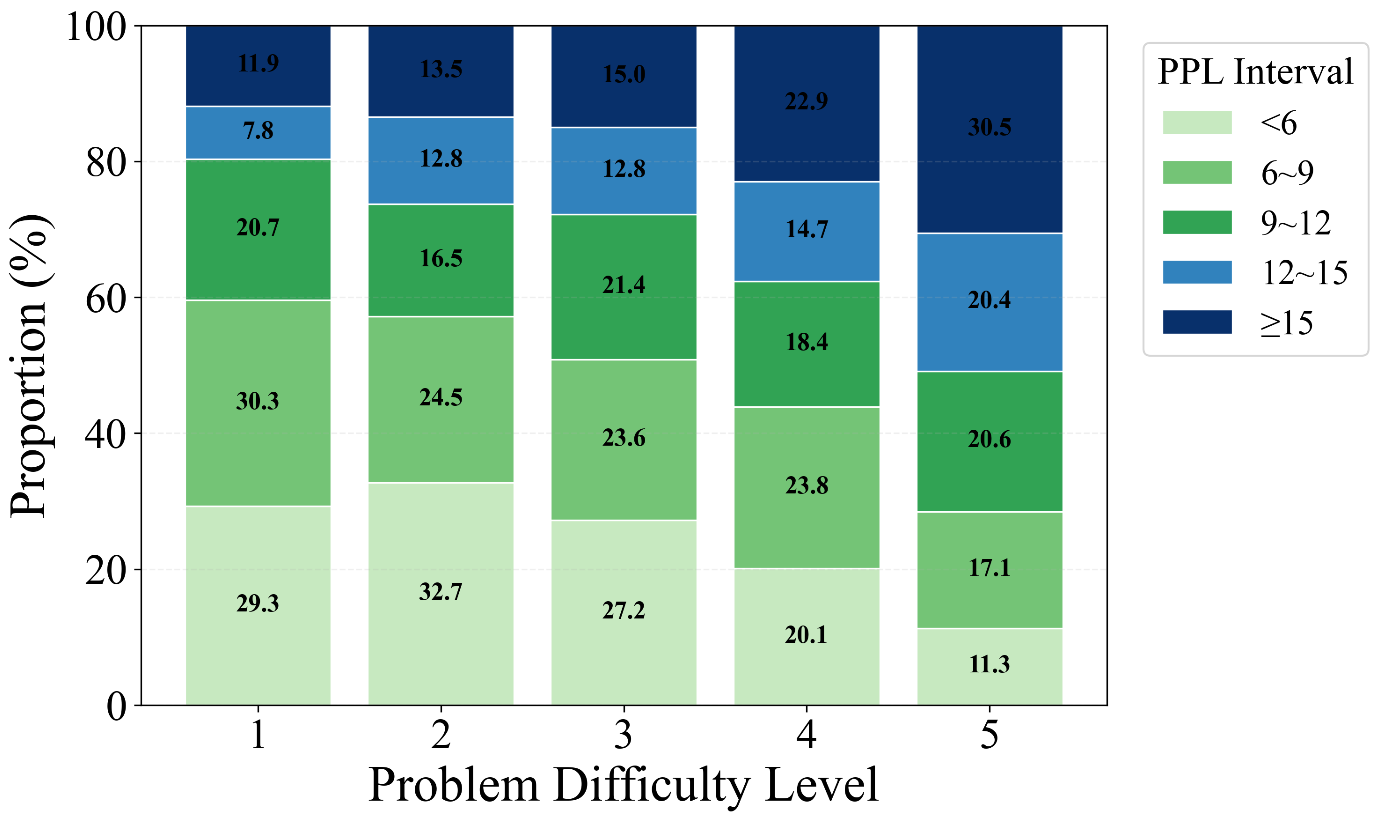}
    \caption{The high PPL rate distribution of different difficulty answers from MATH benchmark, the abscissa is 5 difficulty levels, and the ordinate is the proportion.}
    \label{fig:moti3}
\end{figure}

\subsection{High-PPL Tokens in Distillation Data}

We conduct a difficulty-stratified analysis on the MATH dataset. We adopt Llama3.2-3B-Instruct as the base model $\mathcal{M}_{\theta}$ and build a reference model $\mathcal{M}_{\text{ref}}$ to generate reasoning paths conditioned on golden answers. We compute token-level perplexity from the base model and empirically regard tokens with perplexity > 2.5 as \textit{high-PPL tokens}, which indicate distribution mismatch with the base model. We then evaluate high-PPL token ratios across difficulty levels.

Figure~\ref{fig:moti3} presents segment-level PPL distributions across different difficulties. Simple questions are dominated by low-PPL content, while ultra-high-PPL($\geq$ 15) segments grow consistently as difficulty increases. For Level 1 easy samples, sequences with PPL lower than 9 take the majority, and ultra-high-PPL content accounts for only 11.89\%. By contrast, Level 5 hard samples yield a sharp drop in low-PPL proportion, with ultra-high-PPL segments rising to 30.52\%. These results demonstrate that hard reasoning induces severe distribution drift under reference-guided generation.

Figure~\ref{fig:moti1} further illustrates the correlation between high-PPL ratio and downstream performance. As high-PPL tokens increase, model performance gradually declines. Fine-tuning on easy data brings stable gains, whereas hard-level training drops overall accuracy by 7\%. This indicates that distribution drift impairs knowledge learning and corrupts inherent model representations. Naive self-distillation blindly fits drifted noisy tokens, which becomes the key limitation for complex reasoning.

\begin{figure}[!t]
    \centering
    \includegraphics[width=0.45\textwidth]{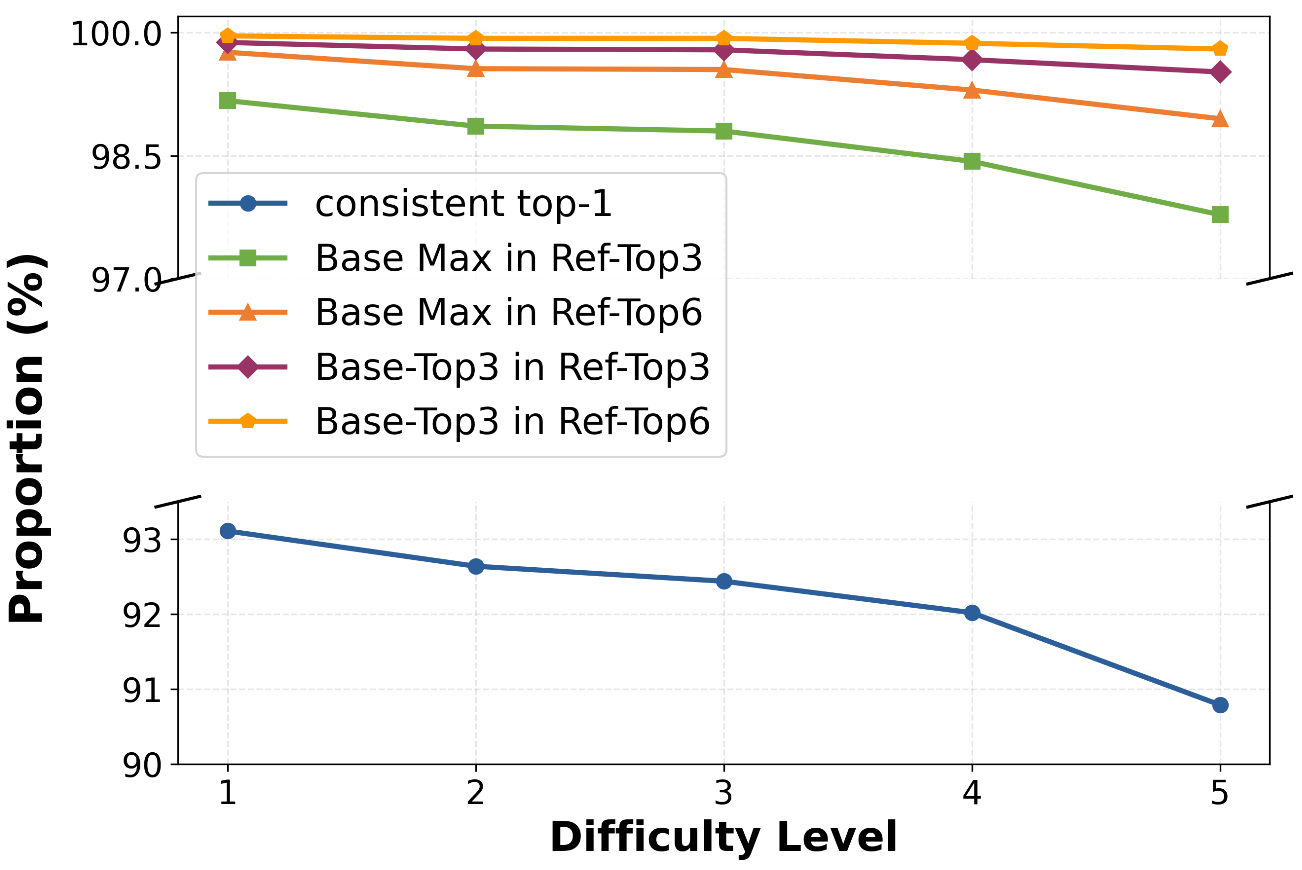}
    \caption{Token-level alignment between $\mathcal{M}_{\theta}$ and $\mathcal{M}_{\text{ref}}$ at different difficulty levels. All metrics are calculated on full sequences sampled from $\mathcal{M}_{\text{ref}}$, including pairwise top-1 token consistency, coverage of $\mathcal{M}_{\theta}$ top-1 token in $\mathcal{M}_{\text{ref}}$ top-3/top-6, and coverage of $\mathcal{M}_{\theta}$ top-3 tokens in $\mathcal{M}_{\text{ref}}$ top-3/top-6.}
    \label{fig:moti4}
\end{figure}

\subsection{Distribution Shift Phenomenon}



To analyze the prediction gap and distribution drift between reference model and base model, we conduct token-level alignment experiments across five difficulty levels. We sample complete sequences from reference model $\mathcal{M}_{\text{ref}}$ and compute token-wise overlap rates with base model $\mathcal{M}_{\theta}$. We measure five statistics: consistent top-1 tokens of both models, the top-1 token of $\mathcal{M}_{\theta}$ covered in $\mathcal{M}_{\text{ref}}$ top-3 and top-6, as well as $\mathcal{M}_{\theta}$ top-3 tokens covered in $\mathcal{M}_{\text{ref}}$ top-3 and top-6.

As shown in Figure~\ref{fig:moti4}, all alignment metrics show a steady downward trend with the growth of reasoning difficulty. The proportion of identical top tokens predicted by two models declines from 93.11\% to 90.79\%, which manifests that high-complexity reasoning enlarges the divergence of the model’s most confident predictions. Although the overall token overlap remains at a high level, the coverage ratio of the base model’s optimal token in reference top-3 candidates gradually decreases from 99.17\% to 97.78\%. Similarly, the overlapping degree of broader base top-3 candidate sets also presents a slow but continuous drop on hard samples.

Such empirical observations provide solid motivation for our DASD method. On the one hand, the extremely high candidate coverage proves that most native high-quality tokens of the base model are naturally included in the reference candidate space. It is feasible to retain the original generation style within the correctness boundary constrained by $\mathcal{M}_{\text{ref}}$. On the other hand, the non-negligible prediction gap on difficult tasks verifies the necessity of fine-grained filtering. Simply following the reference model will inevitably introduce drifted tokens and destroy the base model’s inherent distribution. Therefore, instead of rigid imitation or direct token replacement, our method dynamically balances dual-model outputs via confidence calibration and candidate screening, thereby alleviating distribution drift while maintaining logical correctness for complex mathematical reasoning.

%% file: paper_content/4method.tex
\begin{figure*}[!t]
    \centering
    \includegraphics[width=0.95\textwidth]{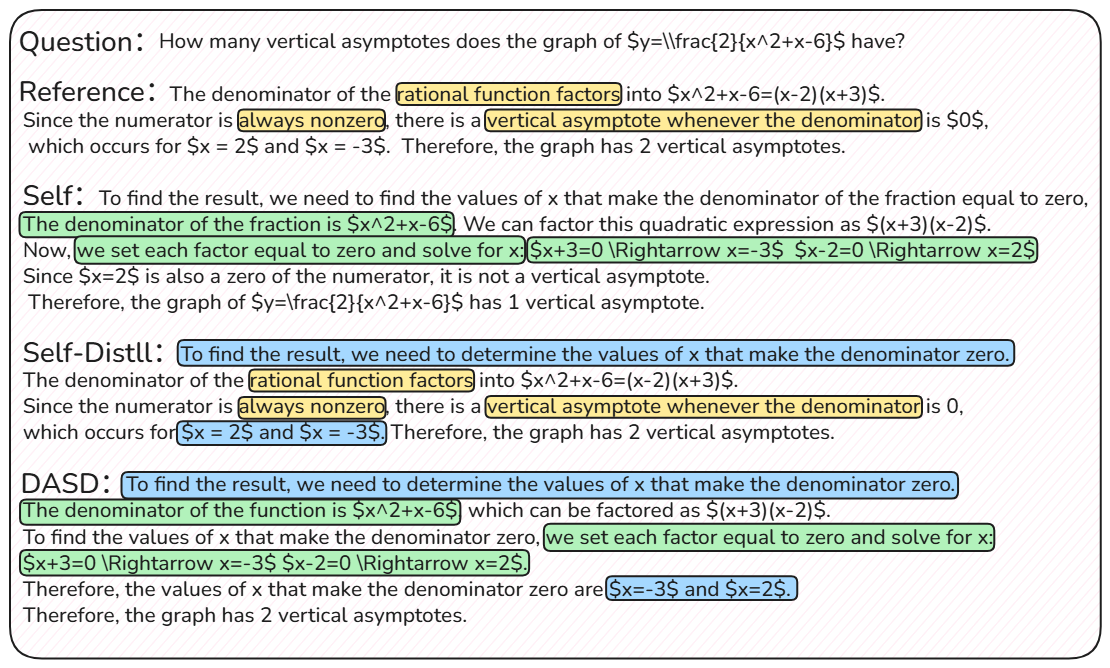}
   \caption{A generation case on the MATH dataset. \textbf{Question}: original problem; \textbf{Reference}: standard solution; \textbf{Self}: base model output; \textbf{Self-Distill}: self-distilled result; \textbf{DASD}: output of our method. Yellow tokens denote mechanical imitation of reference answers that diverge from the base model style. Green tokens retain the inherent linguistic style of the base model, and blue tokens follow reasoning logic consistent with reference-guided outputs.}
    \label{fig:framework}
\end{figure*}

\begin{algorithm}[t]
\caption{Distribution-Aligned Self-Distillation}
\label{alg:dasd}
\begin{algorithmic}[1]
\REQUIRE Base Model $\mathcal{M}_{\theta}$, Reference Model $\mathcal{M}_{ref}$, Dataset $\mathcal{D} = \{(x_i, y_i^*)\}_{i=1}^N$, Confidence Threshold $\tau$, Candidate Number $K$.
\ENSURE Optimized Model $\mathcal{M}_{\theta}$.

\STATE \textbf{Phase 1: Distribution-Aligned Data Generation}
\STATE Initialize aligned dataset $\mathcal{D}_{aligned} \leftarrow \emptyset$ 
\FOR{each sample $(x, y^*)$ in $\mathcal{D}$}
    \STATE Initialize generated sequence $y \leftarrow []$     \FOR{$t = 1$ \TO $T_{max}$}
        \STATE Obtain Top-$K$ candidate tokens $\mathcal{C}_t = \{c_1, \dots, c_K\}$ from $\mathcal{M}_{ref}(x, y, y^*)$         \STATE $selected\_token \leftarrow \text{None}$         \FOR{$c_i$ in $\mathcal{C}_t$}
            \STATE Calculate probability $p = P_{\theta}(c_i | x, y)$             \IF{$p > \tau$}
                \STATE $selected\_token \leftarrow c_i$                 \STATE \textbf{break}
            \ENDIF
        \ENDFOR
        \IF{$selected\_token$ is None}
            \STATE $selected\_token \leftarrow c_1$         \ENDIF
        \STATE Append $selected\_token$ to $y$         \IF{$selected\_token$ is EOS}
            \STATE \textbf{break}
        \ENDIF
    \ENDFOR
    \STATE $\mathcal{D}_{aligned} \leftarrow \mathcal{D}_{aligned} \cup \{(x, y)\}$ \ENDFOR

\STATE \textbf{Phase 2: Model Training}
\STATE Optimize $\mathcal{M}_{\theta}$ on $\mathcal{D}_{aligned}$ using standard Causal Language Modeling loss.
\end{algorithmic}
\end{algorithm}

\section{Method}
In this section, we present Distribution-Aligned Self-Distillation (DASD), a novel self-distillation method to mitigate performance degradation brought by cross-model distribution mismatch. DASD guarantees reasoning correctness by leveraging the base model with golden answer guidance to construct teacher-style reasoning content, and restricts generated content to fit the base model’s inherent output distribution through confidence verification. The resulting distribution-aligned data facilitates effective knowledge transfer while avoiding catastrophic forgetting. The overall procedure of our DASD method is formally described in Algorithm~\ref{alg:dasd}.A concrete generation case comparison is illustrated in Figure~\ref{fig:framework}.

\subsection{Symbol Definition}
Let $\mathcal{M}_{\theta}$ denote the base model to be improved, and $\mathcal{M}_{\text{ref}}$ denote the reference model, which is the same as the base model but guided by golden answers to generate teacher-style reasoning content. The training dataset is $\mathcal{D} = \{(x_i, y_i^*)\}_{i=1}^N$, where $x_i$ is the input and $y_i^*$ is the corresponding golden answer. Let $t$ be the generation time step, $y_{<t}$ be the generated prefix, and $T_{\text{max}}$ be the maximum generation length. Let $K$ be the number of candidate tokens and $\tau$ be the confidence threshold. $\text{EOS}$ denotes the end-of-sequence token.

\subsection{Reference-Guided Candidate Generation}
To guarantee the logical correctness of distilled data, we first generate a high-quality token candidate pool using the reference model. At each generation step $t$, the reference model ($\mathcal{M}_{\text{ref}}$), which is derived from the same base model, computes the conditional probability distribution with access to the golden answer $y^*$:
\begin{equation}
P_{\text{ref}}(\cdot \mid x, y_{<t}, y^*)
\end{equation}
We select the top-$K$ tokens with the highest probabilities to form the candidate set:
\begin{equation}
\mathcal{C}_t = \{c_1, c_2, \dots, c_K\},
\end{equation}
where $c_1$ is the greedy decoding output of $\mathcal{M}_{\text{ref}}$, representing the most reasonable reasoning choice at this step. This step leverages the golden answer to avoid logical fallacies caused by free generation and provides a reliable candidate foundation for subsequent distribution alignment.

\subsection{Dynamic Selection Strategy}
We further design a confidence-verified dynamic selection strategy to balance reasoning correctness and distribution consistency. For each candidate token $c_i$ in the candidate set $\mathcal{C}_t$, we calculate its conditional probability from the base model $\mathcal{M}_{\theta}$, which reflects the native certainty and perplexity level of the model toward each token:
\begin{equation}
p = P_{\theta}(c_i \mid x, y_{<t}).
\end{equation}
This probability acts as the core measurement for distribution alignment. We adopt a two-rule selection mechanism to determine the token to be generated at each step $t$:
\begin{itemize}
    \item Distribution Alignment First: Traverse $\mathcal{C}_t$ and select the first token satisfying $p > \tau$, ensuring outputs conform to the original distribution of the base model.
    \item Knowledge Backup Mechanism: If all K candidates fail the confidence check, we directly choose $c_1$ from $\mathcal{M}_{\text{ref}}$ to maintain reasonable and correct reasoning logic.
\end{itemize}
We repeat this selection process token by token until the EOS token is generated, thus forming a complete and distribution-aligned sequence.

\subsection{Training Objective}
We construct the distribution-aligned dataset $\mathcal{D}_{\text{aligned}}$ through the proposed generation pipeline. Specifically, we filter the generated samples to retain only those with correct answers, which are then used for training the base model. The base model is optimized by standard causal language modeling loss:
\begin{equation}
\mathcal{L}(\theta) = -\sum_{(x, y) \in \mathcal{D}_{\text{aligned}}} \sum_{t=1}^{|y|} \log P_{\theta}(y_t \mid y_{<t}, x; \theta).
\end{equation}

%% file: paper_content/5experiment.tex
\begin{table*}[t]
\centering
\caption{Main Results on Reasoning Benchmarks}
\label{tab:main_results}
\renewcommand{\arraystretch}{1.10}
\small
\begin{tabular*}{0.95\textwidth}{@{\extracolsep{\fill}} lccccc}
\toprule
Model Name & Method & MATH & MBPP & ARC-Challenge & Average \\
\midrule
\multirow{5}{*}{Llama3.2-3B}
& Origin & 0.360 & 0.470 & 0.740 & 0.523 \\
& Self-Distillation & 0.384 & 0.466 & 0.744 & 0.531 \\
& Mask-PPL & 0.388 & 0.490 & 0.762 & 0.547 \\
& Hint-decoding & \textbf{0.418} & \underline{0.500} & \underline{0.770} & \underline{0.563} \\
& DASD & \underline{0.416} & \textbf{0.504} & \textbf{0.780} & \textbf{0.567} \\
\midrule
\multirow{5}{*}{Qwen3-4B}
& Origin & 0.630 & 0.614 & 0.880 & 0.708 \\
& Self-Distillation & 0.670 & 0.532 & 0.888 & 0.697 \\
& Mask-PPL & \underline{0.692} & 0.556 & 0.884 & 0.711 \\
& Hint-decoding & 0.684 & \underline{0.574} & \underline{0.889} & \underline{0.716} \\
& DASD & \textbf{0.702} & \textbf{0.632} & \textbf{0.896} & \textbf{0.743} \\
\midrule
\multirow{5}{*}{Gemma2-2B}
& Origin & 0.220 & 0.370 & 0.700 & 0.430 \\
& Self-Distillation & 0.248 & 0.364 & 0.733 & 0.448 \\
& Mask-PPL & 0.248 & 0.364 & \underline{0.749} & 0.454 \\
& Hint-decoding & \underline{0.262} & \underline{0.370} & 0.745 & \underline{0.459} \\
& DASD & \textbf{0.274} & \textbf{0.380} & \textbf{0.755} & \textbf{0.470} \\
\bottomrule
\end{tabular*}
\end{table*}

\section{Experiments}

\subsection{Experimental Setup}

\textbf{Datasets}
    \label{subsec:datasets}
We conduct experimental validations on three datasets, namely MATH~\cite{hendrycks2021measuring}, MBPP~\cite{austin2021program}, and ARC-Challenge~\cite{clark2018think}. Following the setting in self-distillation research~\cite{yang2024self}, for datasets containing more than 10,000 samples, we randomly select 2,000 samples for fine-tuning to ensure comparable data scale across all datasets.

\textbf{Target Models}
    \label{subsec:target_models}
We evaluate three open-source LLMs with diverse architectures: Llama3.2-3B~\cite{grattafiori2024llama}, Qwen3-4B~\cite{yang2025qwen3}, and Gemma2-2B~\cite{team2024gemma}, all of which are retrieved from Hugging Face.

\textbf{Baseline}
    \label{subsec:baselines}
We compare our proposed method with three baseline methods, including standard self-distillation~\cite{yang2024self}, Mask-PPL~\cite{wu2026mitigating}, and Hint-decoding~\cite{zhang2026towards}, we use the hyperparameters corresponding to their best performance reported in the original papers

\textbf{Evaluation Metrics}
    \label{subsec:metrics}
We perform different strategy-based data processing following the same dataset split, train the original model, and evaluate its performance on the test set. Specifically, we use the ACC score for the MATH and ARC datasets, and the Pass@1 score for the MBPP dataset.

\textbf{Implementation Details}
We adopt LoRA training using the PEFT library, where all methods share the same training parameters: the learning rate is set to 2e-5, the warmup ratio is 0.2, and the target modules are all linear layers. For our DASD method, the confidence threshold is set to 0.2 and the number of candidate tokens is 6.

\subsection{Main Results}
\label{subsec:main_experiments}
Table~\ref{tab:main_results} reports results on Math, MBPP, and ARC-Challenge. Overall, our proposed DASD method achieves superior performance across all three models and three datasets, demonstrating its strong generalization ability. Specifically, for Llama3.2-3B on the MATH dataset, DASD achieves an accuracy of 0.416, which represents a 16\% improvement over the original base model and an 8\% improvement compared to the standard self-distillation method. Notably, on the MBPP dataset, standard self-distillation leads to performance degradation, indicating that models are more susceptible to the influence of golden answers in code generation tasks, which in turn causes severe distribution shift. In contrast, our DASD method effectively alleviates this issue and achieves consistent performance improvements across all evaluated models on MBPP, verifying its effectiveness in mitigating distribution mismatch caused by reference imitation.

\subsection{Ablation Study}

We conduct ablation experiments on the MATH dataset to analyze the effects of the \textbf{confidence threshold}, \textbf{hard token selection strategy}, and \textbf{reference candidate token number}. The results are shown in Table~\ref{tab:ablation}.

First, we explore three confidence thresholds: 0.25, 0.20, and 0.10. As shown in Table~\ref{tab:ablation}, a high threshold of 0.25 obtains an accuracy of 0.406. Strict threshold constraints force the model to retain more native base-model tokens, which harms reasoning correctness. When the threshold is reduced to 0.10, the performance drops slightly to 0.412, since loose thresholds weaken the distribution alignment effect. The threshold of 0.20 achieves the best accuracy of 0.416, striking a favorable balance between reasoning quality and feature distribution consistency.

We further explore the effect of reference candidate size, namely Top-3, Top-6 and Top-9.
A smaller Top-K restricts generation within a narrow reference range, which enhances external knowledge injection but exacerbates distribution inconsistency.
Top-3 adopts strict reference constraints and obtains an accuracy of 0.408.
By contrast, Top-9 relaxes reference restrictions and mitigates distribution drift, yet insufficient logical guidance leads to suboptimal performance of 0.414.
The default Top-6 achieves the optimal accuracy of 0.416, properly balancing reference-based reasoning enhancement and native distribution preservation.

Furthermore, we compare two hard token strategies when all candidate tokens fall below the threshold. The base-mode selects the most confident token from the base model candidates with an accuracy of 0.408. In contrast, the ref-mode adopts the top token from the reference model and reaches 0.416. The results verify that prioritizing correct logic tokens and sufficient knowledge injection via reference guidance is critical for improving mathematical reasoning performance.

\begin{table}[t]
\centering
\caption{Ablation results on the MATH dataset.}
\label{tab:ablation}
\resizebox{0.42\textwidth}{!}{
\renewcommand{\arraystretch}{1.2}
\footnotesize
\begin{tabular}{lc}
\hline
Variant & MATH Acc. \\
\hline
\textbf{Confidence Threshold} & \\
Threshold = 0.25 & 0.406 \\
Threshold = 0.20 & \textbf{0.416} \\
Threshold = 0.10 & 0.412 \\
\hline
\textbf{Candidate Number} & \\
Top-3 & 0.408 \\
Top-6 & \textbf{0.416} \\
Top-9 & 0.414 \\
\hline
\textbf{Hard Token Strategy} & \\
Base-mode & 0.408 \\
Ref-mode & \textbf{0.416} \\
\hline
\end{tabular}
}
\end{table}

\subsection{Analysis}
\subsubsection{Token-Level PPL}
We analyze the average ratio of high-PPL tokens and its discrepancy across samples with different difficulty levels on the MATH dataset, between standard self-distillation and our DASD method in Figure~\ref{tab:ppl_ratio_math}. We observe that DASD significantly reduces the proportion of high-PPL tokens in generated samples, indicating better alignment with the original model distribution. Specifically, the ratio of tokens with PPL greater than 2.5 is 10.59\% in vanilla self-distillation, while DASD reduces it to 4.09\%. Furthermore, the high-PPL ratio generated by DASD remains more consistent across different difficulty levels, with only a 7\% gap between easy and hard samples, compared to 43\% in standard self-distillation. This demonstrates that our distribution-aware filtering mechanism exhibits stronger robustness when facing complex inputs.

\begin{table}[t]
\centering
\caption{High-PPL token ratio (\%) on the MATH dataset under different difficulty levels. Distill presents self-distillatio method}
\resizebox{\linewidth}{!}{
\begin{tabular}{lcc cc}
\toprule
& \multicolumn{2}{c}{PPL $> 2.5$} & \multicolumn{2}{c}{PPL $> 5.0$} \\
Difficulty & Distill & DASD & Distill & DASD \\
\midrule
\underline{Avg.}  & \underline{10.59} & \underline{4.09} & \underline{5.89} & \underline{0.75} \\
Easy   & 9.05  & 4.01 & 4.86 & 0.82 \\
Mid   & 10.19 & 4.03 & 5.65 & 0.70 \\
Hard     & 12.94 & 4.29 & 7.42 & 0.75 \\
\bottomrule
\end{tabular}
}
\vspace{-0.5em}
\label{tab:ppl_ratio_math}
\footnotesize
\end{table}

\begin{figure}[!t]
    \centering
    \includegraphics[width=0.3\textwidth]{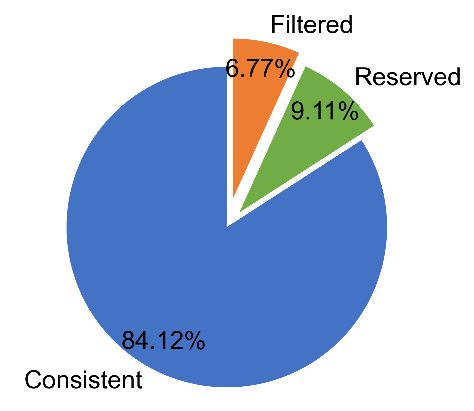}
    \caption{Proportion distribution of three types of token, including Consistent, Reserved, and Filtered tokens.}
    \label{fig:ana1}
\end{figure}

\subsubsection{Token Distribution}
\begin{figure}[!t]
    \centering
    \includegraphics[width=0.4\textwidth]{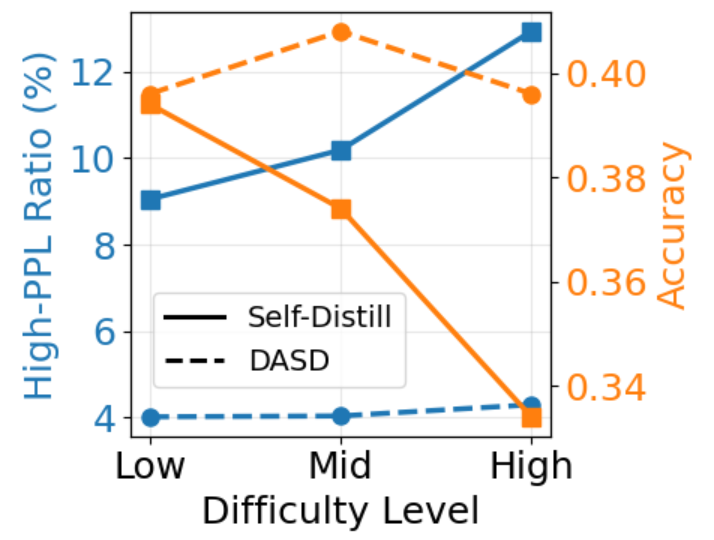}
    \caption{Performance comparison of Self-Distillation and DASD across different difficulty levels. 
    Yellow lines represent accuracy, dashed lines denote the results obtained by DASD method. }
    \label{fig:ana3}
\end{figure}

We carry out training experiments on Llama3.2-3B with MATH data categorized by difficulty levels. As depicted in Figure~\ref{fig:ana3}, we compare the performance of baseline training and DASD-enhanced training across different difficulty tiers. The baseline suffers evident performance fluctuations due to distribution drift, while DASD delivers stable and robust gains at all difficulty levels. Models trained on medium and high-difficulty data outperform those trained on low-difficulty samples, proving that DASD can effectively incorporate valuable reasoning information to promote model optimization. Medium-difficulty training finally achieves the best overall capability, realizing a sound trade-off between external reasoning knowledge infusion and the original distribution characteristics of the model.

\subsubsection{Token Distribution}



We statistically analyze the token selection results of our DASD method on 2000 samples. We focus on five token types in constrained decoding. The consistent token denotes the top-probability output of both the base and reference models. The reserved token refers to valid reference candidates accepted via the confidence threshold. The filtered token emerges when reference candidates fail to meet the threshold and trigger dynamic filtering. The base-selected token is the optimal output selected from the base model after filtering. The hard token enforces the top reference candidate when all reference candidates are unqualified. Only tokens from correctly answered samples are counted to eliminate interference from incorrect reasoning.

Figure~\ref{fig:ana1} illustrates the overall distribution of mainstream token categories. The consistent token occupies an absolutely dominant proportion at 92.54\%, which reveals strong prediction alignment between dual models. The reserved token and filtered token account for 5.25\% and 2.21\% respectively. A small proportion of divergent tokens indicates that the two models maintain stable distribution consistency in most reasoning steps, and the conflict requiring additional constraint adjustment only appears in a few generation positions.

\begin{figure}[!t]
    \centering
    \includegraphics[width=0.40\textwidth]{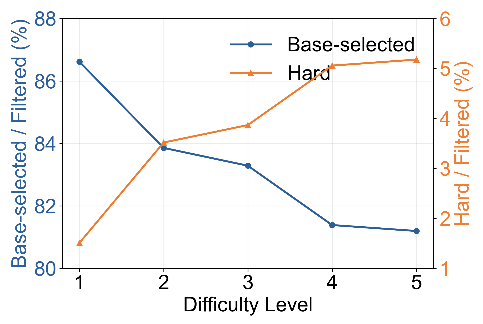}
    \caption{Proportions of base-selected tokens and hard tokens within filtered tokens across different reasoning difficulty levels.}
    \label{fig:ana2}
\end{figure}

Figure~\ref{fig:ana2} further reflects the internal composition variation of filtered tokens across different difficulty levels. The base-selected token always maintains a high proportion within filtered tokens, verifying that our method preferentially retains the optimal output of the base model. As the difficulty level increases, the ratio of base-selected tokens gradually declines, while the proportion of hard tokens increases steadily. This trend demonstrates that complex mathematical reasoning enlarges the distribution discrepancy between dual models. Meanwhile, our constrained decoding framework exhibits great adaptive robustness, which can balance the selection of base model and reference model according to input complexity.

%% file: paper_content/6conclusion.tex
\section{Conclusion}
This paper re-examines self-distillation from the perspective of distribution alignment and proposes DASD, a confidence-aware approach to alleviate distribution drift. By analyzing the negative effects of high-PPL tokens caused by over-reliance on reference answers, we reveal the inherent distribution mismatch issue existing in conventional self-distillation. Experimental results demonstrate that DASD effectively balances external knowledge injection and original distribution preservation, achieving stable and outstanding performance and outperforming traditional self-distillation .

\section*{Limitations}
We adopt dual-path simultaneous decoding to guarantee the correctness of training samples and maintain the distribution consistency of the original model, which achieves favorable performance in offline distillation scenarios. Nevertheless, our sample generation strategy mainly follows greedy token-wise selection, which may lead to insufficient diversity of distilled training data. In future work, we intend to further identify stylistic tokens and generate more diverse training samples while preserving reasoning correctness.

%% file: main.bib
@article{luo2025empirical,
  title={An empirical study of catastrophic forgetting in large language models during continual fine-tuning},
  author={Luo, Yun and Yang, Zhen and Meng, Fandong and Li, Yafu and Zhou, Jie and Zhang, Yue},
  journal={IEEE Transactions on Audio, Speech and Language Processing},
  year={2025},
  publisher={IEEE}
}

@inproceedings{li2024revisiting,
  title={Revisiting catastrophic forgetting in large language model tuning},
  author={Li, Hongyu and Ding, Liang and Fang, Meng and Tao, Dacheng},
  booktitle={Findings of the association for computational linguistics: EMNLP 2024},
  pages={4297--4308},
  year={2024}
}

@inproceedings{yang2024self,
  title={Self-distillation bridges distribution gap in language model fine-tuning},
  author={Yang, Zhaorui and Pang, Tianyu and Feng, Haozhe and Wang, Han and Chen, Wei and Zhu, Minfeng and Liu, Qian},
  booktitle={Proceedings of the 62nd Annual Meeting of the Association for Computational Linguistics (Volume 1: Long Papers)},
  pages={1028--1043},
  year={2024}
}

@article{wu2026mitigating,
  title={Mitigating forgetting in llm fine-tuning via low-perplexity token learning},
  author={Wu, Chao-Chung and Tam, Zhi Rui and Lin, Chieh-Yen and Chen, Yun-Nung Vivian and Sun, Shao-Hua and Lee, Hung-yi},
  journal={Advances in Neural Information Processing Systems},
  volume={38},
  pages={1708--1744},
  year={2026}
}

@article{liu2026profit,
  title={ProFit: Leveraging High-Value Signals in SFT via Probability-Guided Token Selection},
  author={Liu, Tao and Wu, Taiqiang and Yang, Runming and Sun, Shaoning and Wang, Junjie and Yang, Yujiu},
  journal={arXiv preprint arXiv:2601.09195},
  year={2026}
}

@article{zhang2026towards,
  title={Towards On-Policy SFT: Distribution Discriminant Theory and its Applications in LLM Training},
  author={Zhang, Miaosen and Liu, Yishan and Lin, Shuxia and Yang, Xu and Dai, Qi and Luo, Chong and Jiang, Weihao and Hou, Peng and Zeng, Anxiang and Geng, Xin and others},
  journal={arXiv preprint arXiv:2602.12222},
  year={2026}
}

@inproceedings{wang2023self,
  title={Self-instruct: Aligning language models with self-generated instructions},
  author={Wang, Yizhong and Kordi, Yeganeh and Mishra, Swaroop and Liu, Alisa and Smith, Noah A and Khashabi, Daniel and Hajishirzi, Hannaneh},
  booktitle={Proceedings of the 61st annual meeting of the association for computational linguistics (volume 1: long papers)},
  pages={13484--13508},
  year={2023}
}

@article{singh2023beyond,
  title={Beyond human data: Scaling self-training for problem-solving with language models},
  author={Singh, Avi and Co-Reyes, John D and Agarwal, Rishabh and Anand, Ankesh and Patil, Piyush and Garcia, Xavier and Liu, Peter J and Harrison, James and Lee, Jaehoon and Xu, Kelvin and others},
  journal={arXiv preprint arXiv:2312.06585},
  year={2023}
}

@inproceedings{ren2024learn,
  title={I learn better if you speak my language: Understanding the superior performance of fine-tuning large language models with LLM-generated responses},
  author={Ren, Xuan and Wu, Biao and Liu, Lingqiao},
  booktitle={Proceedings of the 2024 Conference on Empirical Methods in Natural Language Processing},
  pages={10225--10245},
  year={2024}
}

@inproceedings{li2025scar,
  title={Scar: Data selection via style consistency-aware response ranking for efficient instruction-tuning of large language models},
  author={Li, Zhuang and Hua, Yuncheng and Vu, Thuy and Zhan, Haolan and Qu, Lizhen and Haffari, Gholamreza},
  booktitle={Proceedings of the 63rd Annual Meeting of the Association for Computational Linguistics (Volume 1: Long Papers)},
  pages={12756--12790},
  year={2025}
}

@inproceedings{ren2025efficiently,
  title={Efficiently Selecting Response Generation Strategies for Synthetic Data Construction by Self-Aligned Perplexity},
  author={Ren, Xuan and Chen, Qi and Liu, Lingqiao},
  booktitle={Findings of the Association for Computational Linguistics: EMNLP 2025},
  pages={11584--11605},
  year={2025}
}

@article{zhang2026embarrassingly,
  title={Embarrassingly simple self-distillation improves code generation},
  author={Zhang, Ruixiang and Bai, Richard He and Zheng, Huangjie and Jaitly, Navdeep and Collobert, Ronan and Zhang, Yizhe},
  journal={arXiv preprint arXiv:2604.01193},
  year={2026}
}

@inproceedings{zelikman2024star,
  title={Star: Self-taught reasoner bootstrapping reasoning with reasoning},
  author={Zelikman, Eric and Wu, Yuhuai and Mu, Jesse and Goodman, Noah D},
  booktitle={Proc. the 36th International Conference on Neural Information Processing Systems},
  volume={1126},
  pages={0--55},
  year={2024}
}

@article{hendrycks2021measuring,
  title={Measuring mathematical problem solving with the math dataset},
  author={Hendrycks, Dan and Burns, Collin and Kadavath, Saurav and Arora, Akul and Basart, Steven and Tang, Eric and Song, Dawn and Steinhardt, Jacob},
  journal={arXiv preprint arXiv:2103.03874},
  year={2021}
}

@article{austin2021program,
  title={Program synthesis with large language models},
  author={Austin, Jacob and Odena, Augustus and Nye, Maxwell and Bosma, Maarten and Michalewski, Henryk and Dohan, David and Jiang, Ellen and Cai, Carrie and Terry, Michael and Le, Quoc and others},
  journal={arXiv preprint arXiv:2108.07732},
  year={2021}
}

@article{clark2018think,
  title={Think you have solved question answering? try arc, the ai2 reasoning challenge},
  author={Clark, Peter and Cowhey, Isaac and Etzioni, Oren and Khot, Tushar and Sabharwal, Ashish and Schoenick, Carissa and Tafjord, Oyvind},
  journal={arXiv preprint arXiv:1803.05457},
  year={2018}
}

@article{grattafiori2024llama,
  title={The llama 3 herd of models},
  author={Grattafiori, Aaron and Dubey, Abhimanyu and Jauhri, Abhinav and Pandey, Abhinav and Kadian, Abhishek and Al-Dahle, Ahmad and Letman, Aiesha and Mathur, Akhil and Schelten, Alan and Vaughan, Alex and others},
  journal={arXiv preprint arXiv:2407.21783},
  year={2024}
}

@article{yang2025qwen3,
  title={Qwen3 technical report},
  author={Yang, An and Li, Anfeng and Yang, Baosong and Zhang, Beichen and Hui, Binyuan and Zheng, Bo and Yu, Bowen and Gao, Chang and Huang, Chengen and Lv, Chenxu and others},
  journal={arXiv preprint arXiv:2505.09388},
  year={2025}
}

@article{team2024gemma,
  title={Gemma 2: Improving open language models at a practical size},
  author={Team, Gemma and Riviere, Morgane and Pathak, Shreya and Sessa, Pier Giuseppe and Hardin, Cassidy and Bhupatiraju, Surya and Hussenot, L{\'e}onard and Mesnard, Thomas and Shahriari, Bobak and Ram{\'e}, Alexandre and others},
  journal={arXiv preprint arXiv:2408.00118},
  year={2024}
}
